\documentclass{article}

\usepackage{arxiv}

\usepackage[utf8]{inputenc} 
\usepackage[T1]{fontenc}    
\usepackage{hyperref}       
\hypersetup{
    colorlinks = True,
    colorlinks = true,
    linkcolor = blue,
    urlcolor  = blue,
    citecolor = blue,
    anchorcolor = blue
}
\usepackage{url}            
\usepackage{booktabs}       
\usepackage{amsfonts}       
\usepackage{nicefrac}       
\usepackage{microtype}      
\usepackage{lipsum}
\usepackage{makecell}
\usepackage{graphicx}
\usepackage{caption}
\graphicspath{ {./images/} }

\title{Recurrent Neural Network for MoonBoard Climbing Route Classification and Generation}

\author{
 Yi-Shiou Duh \\
  Department of Physics\\
  Stanford University\\
  Stanford, CA 94305 \\
  \texttt{allenduh@stanford.edu} \\
   \And
 Ray Chang \\
  Department of Bioengineering\\
  Stanford University\\
  Stanford, CA 94305 \\
  \texttt{jrc612@stanford.edu} \\
}

\begin{document}
\maketitle
\begin{abstract}
Classifying the difficulties of climbing routes and generating new routes are both challenging. Existing machine learning models not only fail to accurately predict a problem’s difficulty, but they are also unable to generate reasonable problems. In this work, we introduced “BetaMove”, a new move preprocessing pipeline we developed, in order to mimic a human climber’s hand sequence. The preprocessed move sequences were then used to train both a route generator and a grade predictor. By preprocessing a MoonBoard problem into a proper move sequence, the accuracy of our grade predictor reaches near human-level performance, and our route generator produces new routes of much better quality compared to previous work. We demonstrated that with BetaMove, we are able to inject human insights into the machine learning problems, and this can be the foundations for future transfer learning on climbing style classification problems.
\end{abstract}


\section{Introduction}
\begin{figure}[bh]
\captionsetup{font=footnotesize}
\centering
\includegraphics[width=0.9\textwidth]{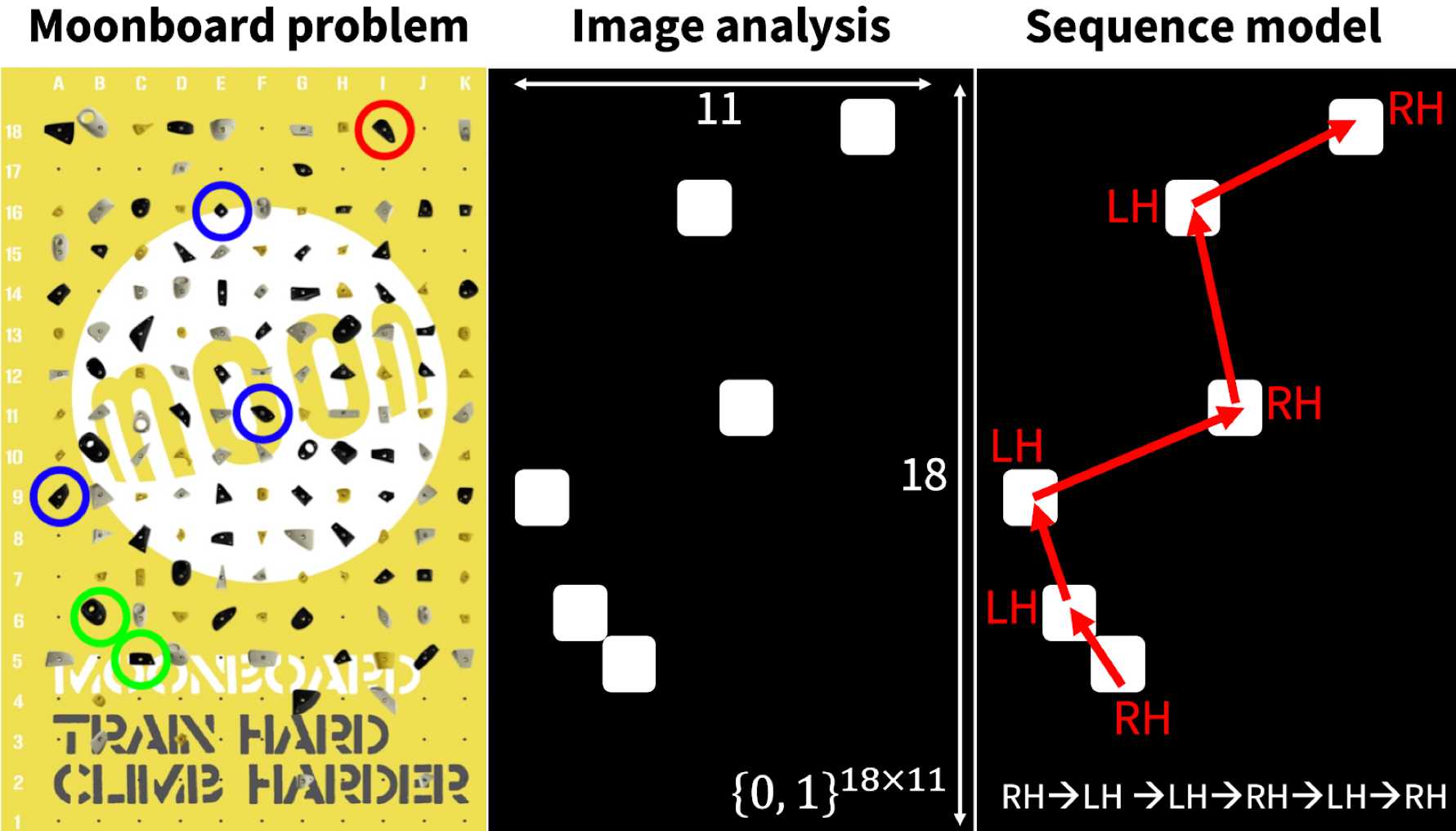}
\caption{Example of a MoonBoard problem and its machine learning model. Previous literature directly analyzed MoonBoard problems as a $\{0, 1\}^{18\times 11}$ matrix. Here we proposed a new sequence model as a more natural representation of a MoonBoard problem. The hand sequence was produced by BetaMove, a new move preprocessing pipeline we developed, in order to mimic a human climber’s hand sequence from the start to the goal.}
\label{fig:fig1}
\end{figure}

The MoonBoard is one of the most effective tools for climbing training. Each MoonBoard is an identical short climbing wall, with 18 rows and 11 columns of holds (left panel of Figure \ref{fig:fig1}). Climbers can only use circled holds to climb a route, also known as a problem, from the start (circled in green) to the goal (circled in red). Each problem has a suggested difficulty, also known as grade, and some high-quality problems are selected into a benchmarked list. The MoonBoard comes with a mobile app, which allows users to access the global database of routes uploaded by the climbing community. The app also has filters that allows users to select specific problems or problems of a certain difficulty.

\begin{figure}[h]
\captionsetup{font=footnotesize}
\centering
\includegraphics[width=0.9\textwidth]{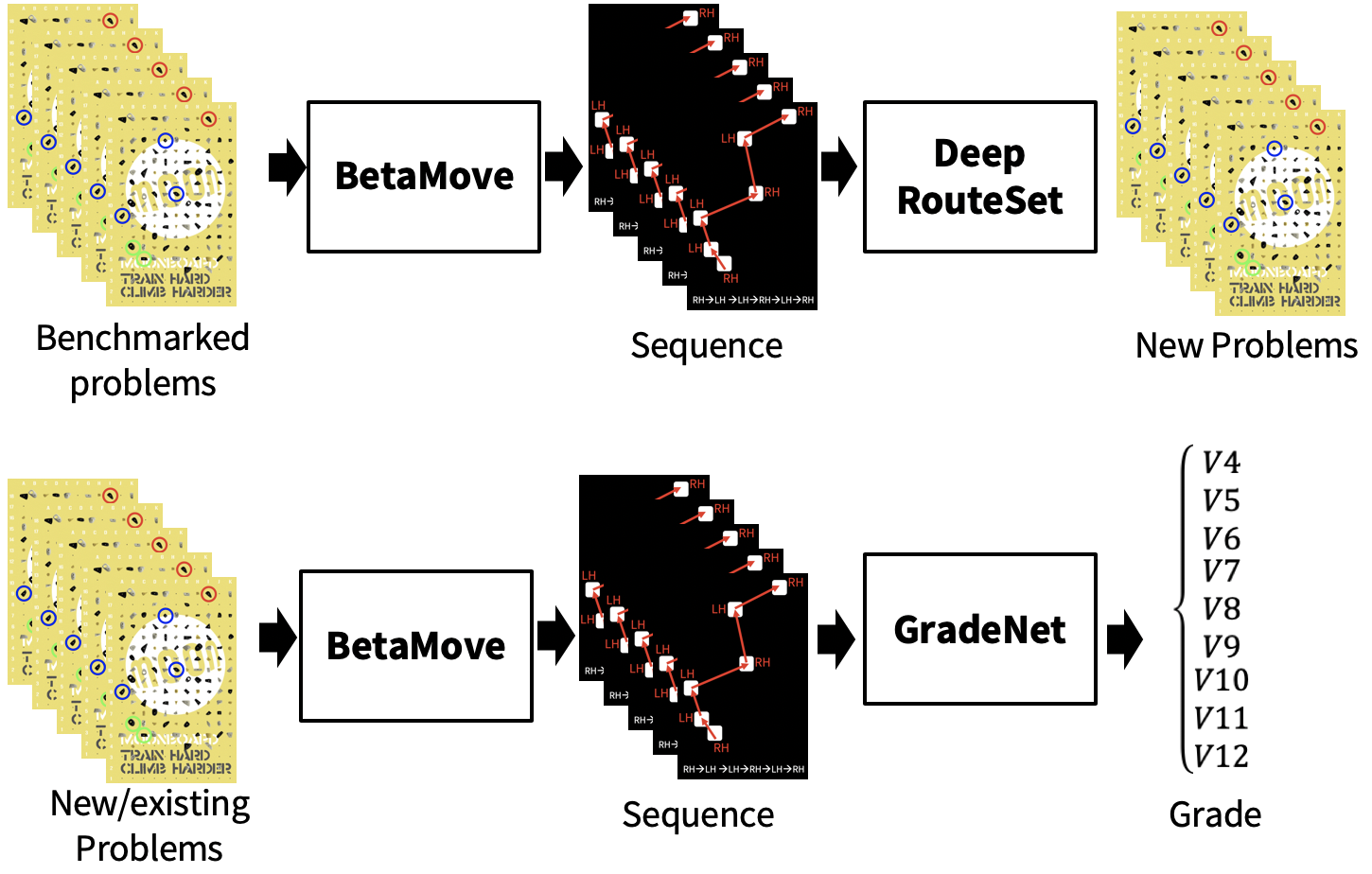}
\caption{The AI pipeline of our project. We first use BetaMove to preprocess MoonBoard problems into a move sequence that is similar to human climber’s prediction. The preprocessed move sequences were then used to train both a route generator, DeepRouteSet, and a grade predictor, GradeNet.}
\label{fig:fig2}
\end{figure}

From a climber’s point of view, it is beneficial to have a route generator that can produce more difficult problems and a grade predictor that can predict their grade. Although machine learning (ML) has been widely applied to sports analysis, rock climbing has not been well-analyzed so far because of the complexity of movements associated with climbing. Current ML models not only fail to accurately predict a problem’s difficulty, but they are also unable to generate “reasonable” problems with natural hand sequences and without redundant holds. We therefore propose to fill this gap using recurrent neural networks (RNN) with the aid of a large database from MoonBoard.

We review several previously published works for climbing grade classification, which all yielded grade predictions lower in accuracy than human predictions. In one example, the authors use convolutional neural networks (CNN) directly on a $\{0, 1\}^{18\times 11}$ matrix to predict the grade of the problems, which resulted in only a 34\% accuracy \cite{Dobles2017}. Tai et al. also encoded the MoonBoard problem as a $\{0, 1\}^{18\times 11}$ matrix, but used graphical convolutional network (GCN) and text embedding technique to obtain an area under curve (AUC) of 0.73 \cite{Tai2020}. Houghton et al. tried many different machine learning frameworks, including RNN \cite{Houghton}. However, in his work, the input sequences were not processed to mimic climbing sequences, which ended up with poor classification accuracy. Kempen used a non-machine-learning method that can only differentiate easy from difficult problems, but the accuracy of their algorithm was still only 64\% \cite{Kempen2019}.

Previous work for artificial intelligence route generation is limited. Houghton et al. created a MoonBoard route generator using long short term memory  (LSTM) \cite{Houghton}. Their training set input sequences are not preprocessed to mimic climbing sequence, so \href{https://ahoughton.com/moon}{their generated problems} include  redundant holds and strange move sequence. For the non-machine-learning approach, Phillips et al. developed an automatic route setter, “StrangeBeta”, based on the mathematical characteristics of a strange attractor \cite{Phillips2012}. However, they described the generated routes with a special language called Climbing route CRDL, which can be ambiguous even to climbing experts, and their system is not applied to MoonBoard. Previous work literature applied end-to-end, non-sequential models directly to unprocessed MoonBoard problems, which and we summarized their approaches in Table \ref{tab:review}. 

\captionsetup{font=footnotesize}
\begin{table}[h!]
\scriptsize
  \begin{center}
    \caption{The summary of previous grade predictors.}
    \label{tab:review}
    \begin{tabular}{|c|c|c|c|c|c|c|}
    \hline
       & \textbf{Ref \cite{Dobles2017}} & \textbf{Ref \cite{Tai2020}}& \textbf{Ref \cite{Kempen2019}} & \textbf{Ref \cite{Houghton}} & \textbf{Ref \cite{Houghton}} & \textbf{Ref \cite{Houghton}} \\
      \hline
      \textbf{Algorithm} & CNN & \makecell{GCN+text\\embedding} & \makecell{CRDL +\\DE-CTW} & LSTM & MLP & \makecell{random\\forest}\\
      \hline
      \textbf{Performance} & \makecell{Accuracy\\34\%} & \makecell{AUC\\0.73} & \makecell{Differentiate\\hard vs easy\\64\%} & \makecell{Accuracy\\29.9\%} & \makecell{Accuracy\\35.6\%} & \makecell{Accuracy\\16.5\%}\\
      \hline
    \end{tabular}
  \end{center}
\end{table}

In our opinion, MoonBoard problems are more similar to an natural language processing (NLP) problem than a computer vision problem due to two major reasons: 1)With graphic representation, the $\{0, 1\}^{18\times 11}$ matrix used in CNN is too sparse and 2)Climbers follow a physically reasonable sequence to climb up, from the start hold to the goal, and a sequential model would best represent that process. We therefore decided to implement RNN with proper preprocessing on the climbing route grading and route generation problem. 

In this paper, we demonstrated an improved version of the grade predictor - “GradeNet” - using a new move preprocessing pipeline “BetaMove.” This approach allows us to outperform other existing classifiers in literature and even reach human-level performance. This advance shows that move sequences generated by BetaMove is a more natural representation compared to sequences generated by other models. Based on this new sequence input, we then built a route generator “DeepRouteSet”. This work not only pioneers the RNN architecture for climbing route classification, but also allows future work to apply transfer learning to learn style-labeled data. The link to our Github repository can be found here: \href{https://github.com/jrchang612/MoonBoardRNN}{https://github.com/jrchang612/MoonBoardRNN}, and a demo website showing our generated problems can be found here: \href{https://jrchang612.github.io/MoonBoardRNN/website/}{https://jrchang612.github.io/MoonBoardRNN/website/}.

\section{Methods}
As shown in Figure \ref{fig:fig2}, the overall pipeline of our project started with BetaMove, which preprocesses the image of a MoonBoard problem into a move sequence. BetaMove simulates a variety of climbing sequences and computes a success score, just like human climbers trying out climbing sequences to find out the easiest one. Each sequence is composed of many moves, in which each move’s success score can be parameterized by the relative distance between holds and the difficulty scale of each hold. We then used beam search algorithm (with beam size = 8) to find the best sequence. The parameters of BetaMove were tuned to match the 20 dev-set hand sequence predicted by a climbing expert, and we evaluated the performance of the model by comparing the predicted move sequences of BetaMove on another 20 MoonBoard problems with the ones predicted by a climbing expert. The prediction of BetaMove exactly matched the move sequences predicted by a climbing expert in 95\% of the test problems (19/20). This physically meaningful move sequence injects human’s insight to improve our input structure.

\begin{figure}[h]
\captionsetup{font=footnotesize}
\centering
\includegraphics[width=1\textwidth]{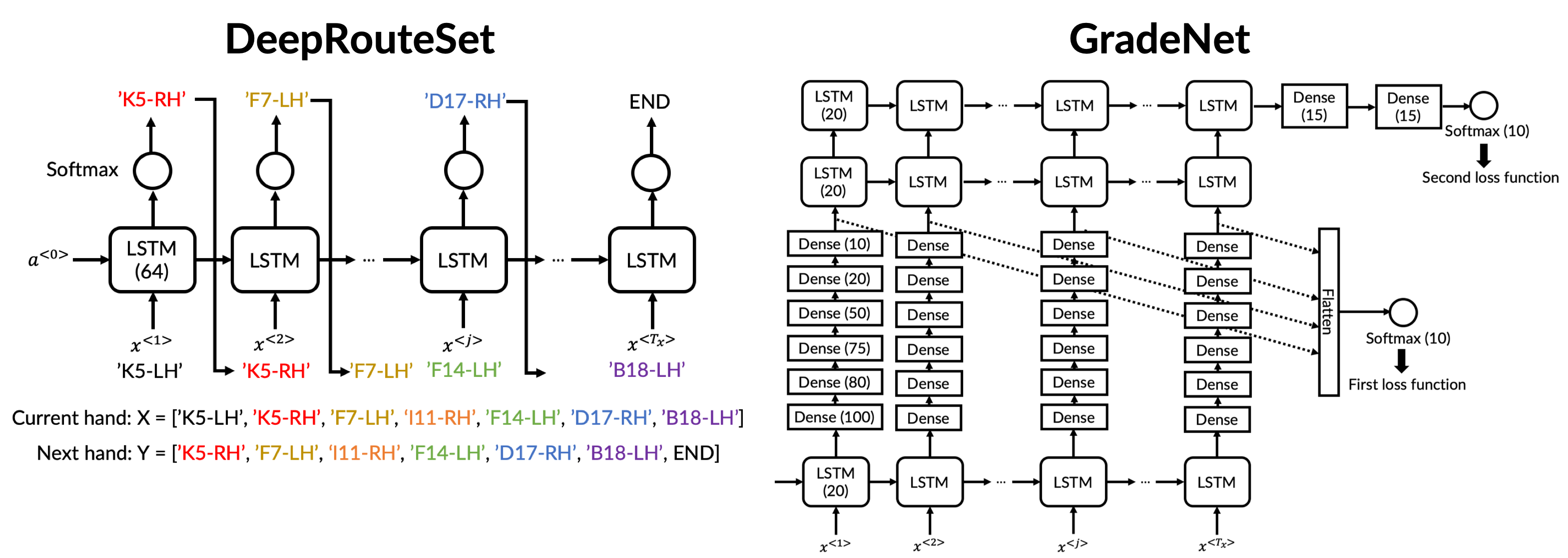}
\caption{Model architecture of DeepRouteSet and GradeNet. DeepRouteSet is a route generation model which is trained with BetaMove-preprocessed sequences. GradeNet is a sentiment classification model trained with the mapping between BetaMove-preprocessed sequences and their corresponding grades. The trained GradeNet can predict the grade of both existing and generated problems. The number in each box indicates the number of neurons in each layer.}
\label{fig:fig3}
\end{figure}

Both our route generator, DeepRouteSet, and our grade predictor, GradeNet, were trained using the sequence data  generated by BetaMove. Similar to music generation, DeepRouteSet learns the pattern between adjacent moves  from existing MoonBoard problems, and is able to generate new problems using these patterns. In GradeNet, we implemented 2 novel techniques. First of all, the input of GradeNet is a move sequence produced by BetaMove, and each move is embedded into a 22-dimensional vector. Secondly, to grade a MoonBoard problem, both hidden layer prediction and overall grade prediction need to be considered at the same time. This spirit was embodied in the two-stage architecture of GradeNet, as shown in the right panel of Figure \ref{fig:fig3}. In the first stage, the embedded input sequences pass through the LSTM layer, followed by 6 dense layers. The output sequence of the 6th dense layer is combined and flattened for the first grade prediction. In the second stage, the output of the 6th dense layer was fed into another two LSTM, followed by 2 dense layers for another grade prediction. The final loss function was the sum of the two categorical cross entropy. Batch normalization, dropout, L2 regularization, max pooling, and average pooling were all tested, but none of them shows significant improvement. The training history of GradeNet was described in Appendix.

\section{Dataset and Features}
\href{https://www.moonboard.com/}{MoonBoard's official website} hosts a database of problems uploaded by the global climbing community. We scraped 30634 MoonBoard problems using a modified Selenium scraping script based on gestalt-howard’s Github project \href{https://github.com/gestalt-howard/moonGen}{moonGen}. Because everyone can upload problems and give it a grade, additional preprocessing is required to filter out problems from amateur users. First, to reduce the ambiguity of grade scale, we chose to unify the grade scale with Hueco scale grading system (scale from V4 to V14, and is more prevalent in Asia and in the US) instead of the Font scale (scale from 6B to 8B+ and is more prevalent in Europe) because both 6C and 6C+ in the Font scale map to V5 in the Heuco scale, and both 7B and 7B+ map to V8. Second, we excluded all V14 problems (24 problems) because many of them are clearly mislabeled with very easy moves. Third, we exclude some problems we found in the error analysis: 1 problem has mislabeled start holds, and 6 problems have unrealistically large numbers of holds. Finally, we removed all problems without repeats, which means no other users have verified the route and usually indicates low quality and questionable grade. The final grade distribution is shown in Appendix Figure \ref{fig:figA3}. The remaining 25096 problems were divided into training, dev, and test set, with 20157, 2442, and 2497 problems, respectively. The distribution is highly skewed toward easy problems, and we use weighted training to counteract this imbalance. BetaMove preprocessed the gleaned dataset into a move sequence for all further training and testing.

\section{Experiments, Results, and Discussion}
To evaluate the grade prediction accuracy of GradeNet, we estimated human-level performance. As described in detail in Appendix A1, the human-level accuracy of exact match is only 45\%. This is because it is difficult to assign a discrete grade from a continuous spectrum of climbing difficulties. If you allow the predicted grade to be off by 1 grade, the accuracy increases to about 87.5\%.

\begin{table}[h!]
\captionsetup{font=footnotesize}
\scriptsize
  \begin{center}
    \caption{
    Performance of GradeNet and previous classifiers. The performance of GradeNet is close to human level performance (HLP). Results from Ref\cite{Dobles2017}, Ref\cite{Tai2020}, and Ref\cite{Houghton} are listed for comparison. In Ref\cite{Houghton}, the author tried LSTM, MLP, and random forest algorithm, and here we only reported their data from MLP as it had the best performance.}
    \label{tab:performance}
    \begin{tabular}{|c|c|c|c|c|c|c|c|c|}
    \hline
       & \textbf{HLP} & \makecell{\textbf{GradeNet}\\\textbf{Training}} & \makecell{\textbf{GradeNet}\\\textbf{Dev set}} & \makecell{\textbf{GradeNet}\\\textbf{Test set}} & \makecell{\textbf{Naive RNN}\\\textbf{Dev set}} & \makecell{\textbf{Ref\cite{Dobles2017}}\\\textbf{CNN}} & \makecell{\textbf{Ref\cite{Tai2020}}\\\textbf{Graph NN}} &
       \makecell{\textbf{Ref\cite{Houghton}}\\\textbf{MLP}}\\
      \hline
      \textbf{Accuracy} & 45.0\% & 64.3\% & 47.5\% & 46.7\% & 34.7\% & 34.0\% & \makecell{Not\\reported} & 35.6\%\\
      \hline
      \makecell{\textbf{$\pm$1}\\\textbf{accuracy}} & 87.5\% & 91.3\% & 84.8\% & 84.7\% & \makecell{Not\\evaluated} & \makecell{Not\\reported} & \makecell{Not\\reported} & 74.5\%\\
      \hline
      \textbf{F1 score} & - & 0.506 & 0.242 & 0.255 & 0.165 &\makecell{Not\\reported}& 0.310 & \makecell{Not\\reported}\\
      \hline
      \textbf{AUC} & - & 0.898 & 0.764 & 0.773 & \makecell{Not\\evaluated} & \makecell{Not\\reported} & 0.73 & \makecell{Not\\reported}\\
      \hline
    \end{tabular}
  \end{center}
\end{table}

The performance of GradeNet, in comparison with existing MoonBoard grade predictors, are summarized in Table \ref{tab:performance}. GradeNet not only outperformed other classifiers including CNN\cite{Dobles2017}, MLP\cite{Houghton}, Graphical neural network\cite{Tai2020}, but also reached human-level performance. GradeNet also surpassed a naive RNN, in which the hold lists of the problems were directly fed as input without the preprocessing of BetaMove. This indicates that the improvement primarily comes from the injection of human insight through the BetaMove-preprocessed move sequence.

\begin{figure}[h]
\captionsetup{font=footnotesize}
\centering
\includegraphics[width=0.9\textwidth]{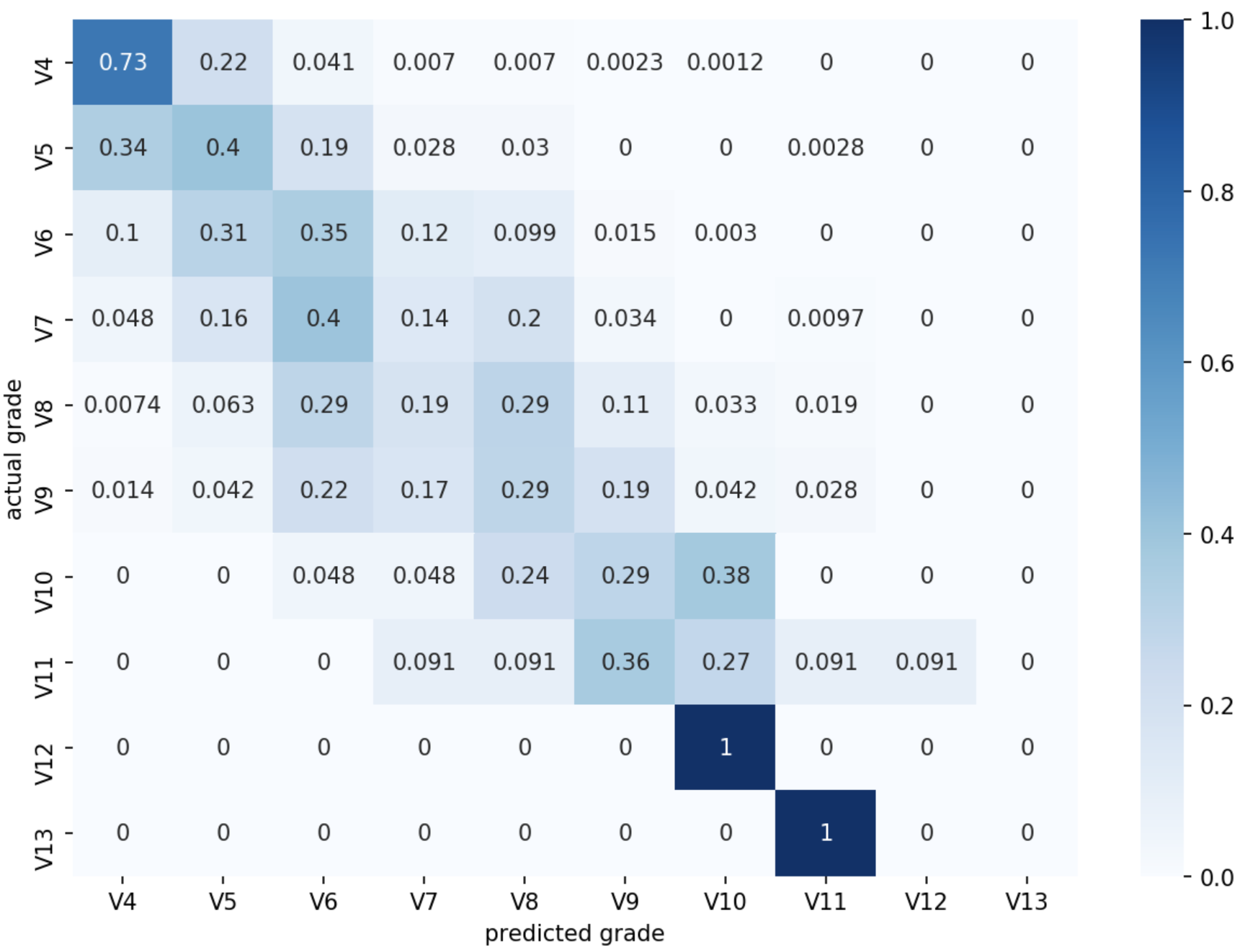}
\caption{The confusion matrix of GradeNet. The predicted grade and actual grade distributed along the main diagonal. For problems easier than V8, GradeNet performs well. For harder problems, GradeNet underestimates their difficulty.}
\label{fig:fig4}
\end{figure}

The confusion matrix in Figure \ref{fig:fig4} shows that the prediction of GradeNet mostly agrees with the actual grade within $\pm$1 error. For the majority of problems (V4-V7), the confusion matrix is symmetric. We believe that the deviation comes from the subjective nature of grading, and the difficulties to categorize problems with difficulties between two grades. For problems harder than V8, however, our model underestimated their grade. This is probably because in those difficult problems, there are only 1 or 2 crux moves while most other moves are around V8. For instance, a V11 problem may only have one V11 crux move. We therefore expect an attention model can better tackle this challenge.

\begin{table}[h!]
\captionsetup{font=footnotesize}
\scriptsize
  \begin{center}
    \caption{Performance of DeepRouteSet, in comparison with the route generator reported in Ref\cite{Houghton}}
    \label{tab:routeset}
    \begin{tabular}{|c|c|c|c|c|c|c|c|}
    \hline
       & \makecell{\textbf{total}\\\textbf{problems}\\\textbf{evaluated}} & \makecell{\textbf{redundant/}\\\textbf{ holds}} & \makecell{\textbf{weird sequence/}\\\textbf{ugly moves/}\\\textbf{lack of flow}} &  
       \makecell{\textbf{reasonable}\\\textbf{problems}} & \makecell{\textbf{high quality}\\\textbf{problems}} \\
      \hline
      \makecell{\textbf{Latest}\\\textbf{MoonBoard}\\\textbf{problems}} & 50 & 10\% & 6\% &  84\% & 60\%\\
      \hline
      \makecell{\textbf{Deep}\\\textbf{RouteSet}} & 40 & 0\% & 5\% &  95\% & 80\%\\
      \hline
      \makecell{\textbf{Model}\\\textbf{Ref\cite{Houghton}}} & 40 & 35\% & 35\% &  40\% & 20\% \\
      \hline
    \end{tabular}
  \end{center}
\end{table}

For our route generator, DeepRouteSet, we asked a climbing expert to evaluate the quality of generated problems by 2 criteria: (1) Is this problem decent/reasonable? A reasonable problem should not have obvious redundant/unused holds, weird sequences that can cause injury, ugly moves, or the coexistence of very easy and very hard moves. (2) Is this problem a high-quality one that can be a candidate for benchmark? High quality problems usually have a natural climbing flow, guidance for climbers to apply a natural posture, and consistent difficulties of moves. The comparison of our model with the route generator reported by Houghton \cite{Houghton} and the latest 50 MoonBoard problems is summarized in Table \ref{tab:routeset}. The differences in the quality is striking. Problems generated by DeepRouteSet showed much less poor qualities compared to the existing model (examples shown in Figure \ref{fig:figA4}). Furthermore, DeepRouteSet produced 80\% of high quality problems compared with  20\% in Ref\cite{Houghton}. Note that in Ref\cite{Houghton}, LSTM was also used for route generation. This striking difference indicates that the human insight from BetaMove can help improve the quality of a route generator.

We have shown that with our new pipeline, we greatly improved the performance of a grade classifier and route generator for climbing problems. In the future, we hope to use the same framework to generate a climbing style classifier. From a climber's point of view, it would be great if each problem can come with a tag which indicates the crux of that problem or the specific moves that might be involved. This would allow climbers to easily find pertinent problems to train a specific climbing skill. However, currently we are not able to collect enough labeled data to train the model. We believe with more style-labeled data, we can apply transfer learning using the pretrained weights from GradeNet.

\section{Conclusion}
In conclusion, we demonstrated that with BetaMove, we are able to inject human insights into the machine learning problems. By preprocessing a MoonBoard problem into a proper move sequence, the accuracy of our grade predictor, GradeNet, reaches near human-level performance. In addition, our strategy also optimizes our route generator, DeepRouteSet. We believe our new processing pipeline can be the foundations for future work on climbing route problems.
This work not only pioneers the RNN architecture for climbing route classification, but also allows future work to apply transfer learning to learn style-labeled data.

\section{Contributions}
RC adapted the source code from gestalt-howard’s Github project \href{https://github.com/gestalt-howard/moonGen}{moonGen} to scrape the MoonBoard data and organized the dataset. YSD developed BetaMove and analyzed the performance of “BetaMove”. RC developed the GradeNet and analyzed the performance of GradeNet. YSD performed the analysis of human-level performance. YSD adapted the source code from \href{https://www.coursera.org/learn/nlp-sequence-models}{a Coursera problem exercise} “Improvise a Jazz Solo with an LSTM Network” and built DeepRouteSet. YSD collected the labeled data for style analysis from himself and his climber friends. RC modified the source code from Andrew Houghton’s Github project \href{https://github.com/andrew-houghton/moon-board-climbing}{moon-board-climbing} to set up the website. RC maintained the GitHub repository. YSD and RC contributed equally to the writing.

\section{Acknowledgement}
The authors appreciated the advice from Colin Wei during the early phase of this project. The authors appreciated the permission from Howard (Cheng-Hao) Tai for using his scraping code. The authors appreciated the permission from Andrew Houghton for using his website code. The authors acknowledged the help from Zhi-Xuan Jian for evaluating the difficulties to grab holds and providing the estimate of human-level performance. The authors acknowledged the help from Ming-Shu Gan in filming and editing climbing videos and evaluated the estimate of human-level performance. The author acknowledged several climbers for providing their video and providing evaluation on generated MoonBoard problems: Maya Madere, Jiun-Jie Tzeng, Hung-Ying Lee, Keita Watabe, Chia-Hsiang Lin, Joey Wen, Jeff Lau, Hoseko Lee, Nathaniel Coleman. The authors acknowledged insightful discussion with Lucien Lo, Yi-Ting Duh, Liting Xu. The authors thanked the help from people who take part in style surveys. The authors acknowledge the help from Hume Center for Writing and Speaking on the manuscript.

\section{Conflict of Interest}
The authors declare no conflict of interests with any existing enterprises. This work is not funded or supported by MoonBoard Company.

\bibliographystyle{unsrt}  
\bibliography{references}  

\begin{thebibliography}{1}

\bibitem{Dobles2017}
Alejandro Dobles, Juan~Carlos Sarmiento, and Peter Satterthwaite.
\newblock {Machine Learning Methods for Climbing Route Classification}, 2017.

\bibitem{Tai2020}
Cheng-Hao Tai, Aaron Wu, and Rafael Hinojosa.
\newblock {Graph Neural Networks in Classifying Rock Climbing Difficulties},
  2020.

\bibitem{Houghton}
Andrew Houghton, Ryan Mann, and Jemma Herbert.
\newblock Moon board climbing.
\newblock \url{https://github.com/andrew-houghton/moon-board-climbing}, 2020.

\bibitem{Kempen2019}
Lindsay Kempen.
\newblock {A fair grade : assessing difficulty of climbing routes through
  machine learning}, 2019.

\bibitem{Phillips2012}
C.~Phillips, L.~Becker, and E.~Bradley.
\newblock {Strange beta: An assistance system for indoor rock climbing route
  setting}.
\newblock {\em Chaos}, 2012.

\end{thebibliography}

\appendix
\counterwithin{figure}{section}
\counterwithin{table}{section}
\section{Appendix}

\textbf{Appendix A1: Human-level Performance}

To help us evaluate the performance of GradeNet, we first estimated human-level performance on the MoonBoard grading problem. In the previous study \cite{Dobles2017}, how much the user-rated grade on MoonBoard app matches the grade provided by route-setters is used as human-level performance. However, we don’t think this is a fair comparison because climbers usually assign grades after they have climbed the problem, and there is a climbing convention to follow the original grade unless there is significant error as a respect to the route setter’s effort.

To combat this problem and make a fair estimation of human-level performance, we instead asked 3 climbing experts to blindly estimate 40 climbing problem’s grades without actually climbing it. As shown in Table \ref{tab:HLP}, it is very difficult even for climbing experts to determine the grade without actually climbing it. However, if you tolerate one grade off from the predicted grade, the accuracy can be up to 87.5\%. From the feedback of our climbing experts, there are several reasons for their low accuracy:
(1) It is very difficult to grade a problem without actually climbing it.
(2) Climbers of different body types are more suitable for different problems.
(3) When a climber believes that the problem falls between 2 grades, they don’t know which grade they should assign.

\begin{table}[h!]
\captionsetup{font=footnotesize}
\small
  \begin{center}
    \caption{Human level performance of estimating the grade of a MoonBoard problem without actually climbing it.}
    \label{tab:HLP}
    \begin{tabular}{l|c|c} 
       & \textbf{Exactly meet with the grade} & \textbf{Allow one grade difference}\\
      \hline
      \textbf{Climbing expert 1} & 47.6\% & 82.5\% \\
      \hline
      \textbf{Climbing expert 1 (second try)} & 30\% & 77.5\% \\
      \hline
      \textbf{Climbing expert 2} & 42.5\% & 87.5\% \\
      \hline
      \textbf{Climbing expert 3} & 45\% & 87.5\% \\
      \hline
      \hline
      \textbf{Estimated HLP} & 45.0\% & 87.5\% \\
    \end{tabular}
  \end{center}
\end{table}

\textbf{Appendix A2: The training of GradeNet}

\begin{figure}[h]
\captionsetup{font=footnotesize}
\centering
\includegraphics[width=0.8\textwidth]{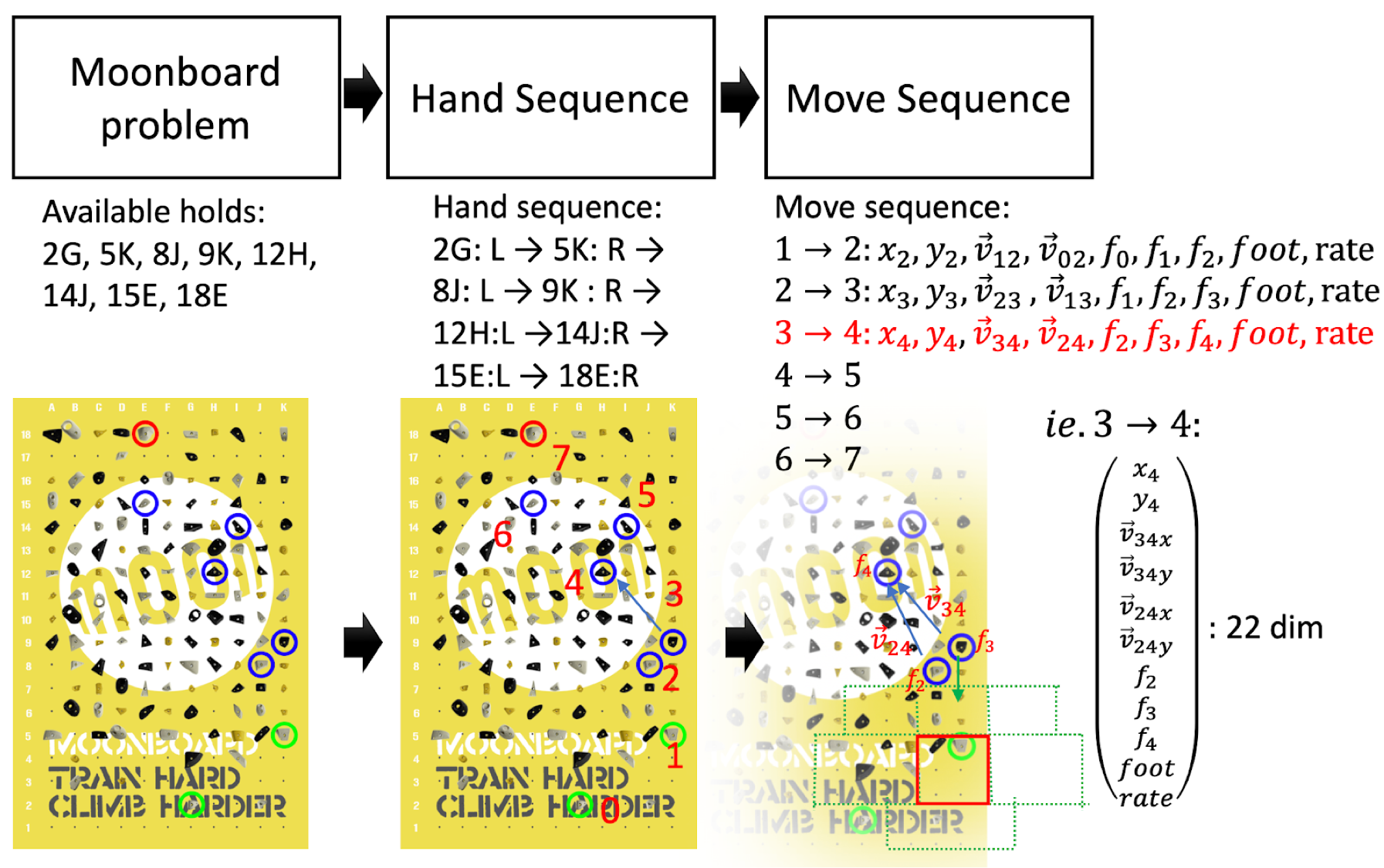}
\caption{Distribution of V-grade of the MoonBoard dataset.}
\label{fig:figA1}
\end{figure}

The input of GradeNet is a move sequence produced by BetaMove. Each move is embedded into a 22-dimensional vector. For example, as shown in Figure \ref{fig:figA1}, BetaMove embedded a single move from hold 3 to hold 4 into a 22- dimensional vector that will be used in GradeNet. This vector includes the target hold's position $(x_4, y_4)$, the relative distance to the previous 2 holds $(v_{34x}, v_{34y}, v_{24x}, v_{24y})$, the difficulty scales of all three holds $(f_2, f_3, f_4)$, the placement of feet, and the estimated success scores of each move. 

\begin{figure}[h]
\captionsetup{font=footnotesize}
\centering
\includegraphics[width=0.9\textwidth]{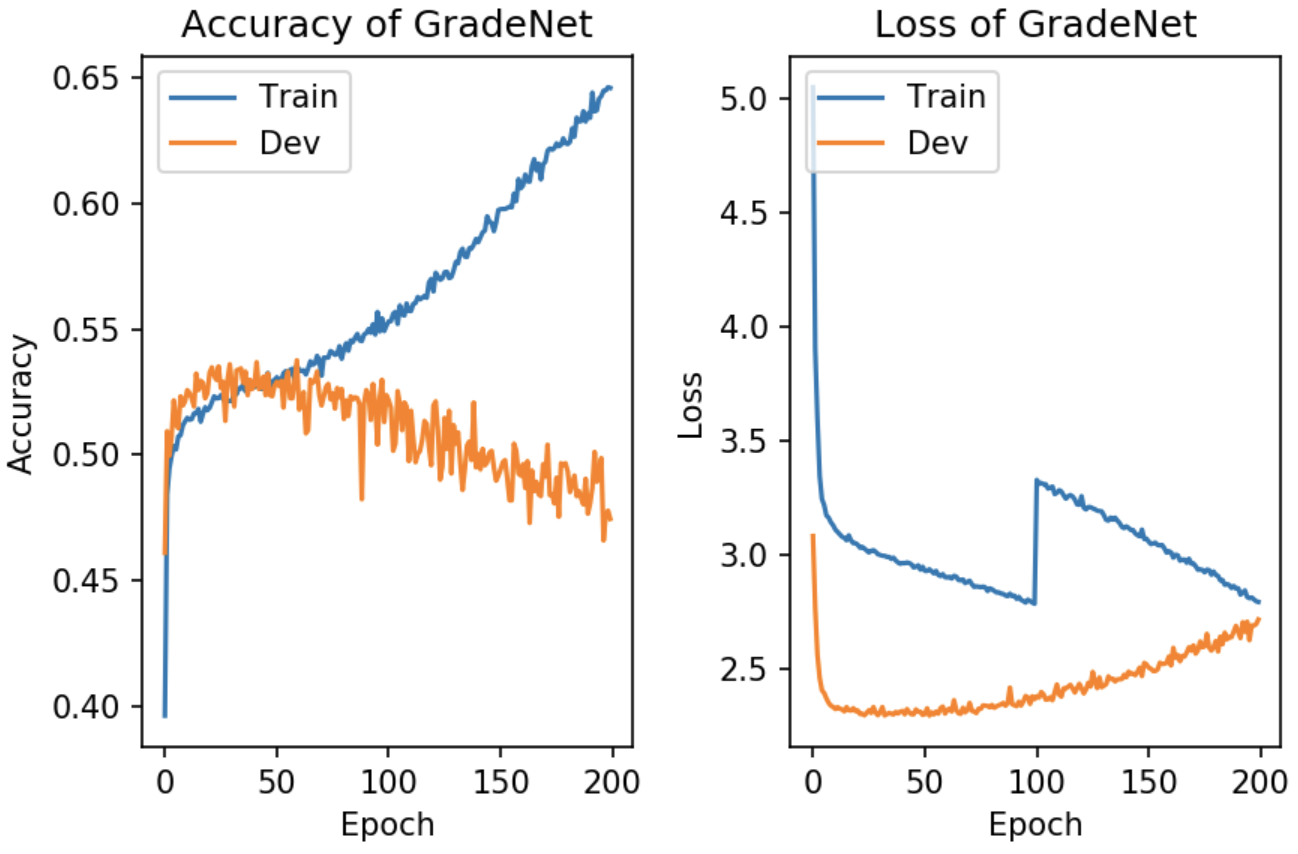}
\caption{The 22-dimensional embedding of BetaMove.}
\label{fig:figA2}
\end{figure}

During the training process (Figure \ref{fig:figA2}), the input was weighted to combat the problem of uneven input class distribution (the weights were adjusted after training for 100 epochs). After the training for 200 epochs, the training results of the model are summarized in Table 1 of the main text.

\begin{figure}[h]
\captionsetup{font=footnotesize}
\centering
\includegraphics[width=0.9\textwidth]{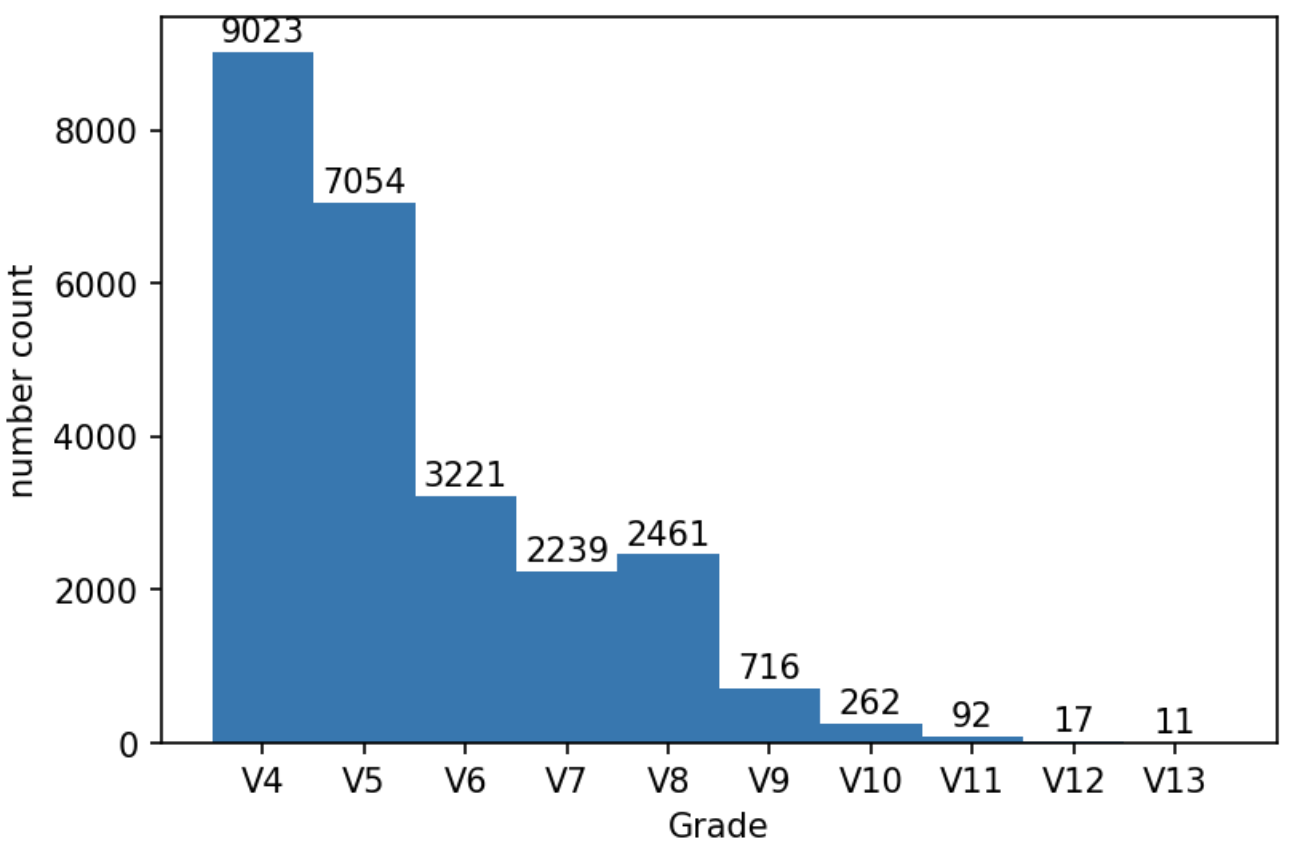}
\caption{Training curve of GradeNet. At Epoch 100, the class weight was adjusted, so the loss function of the training set suddenly increased.}
\label{fig:figA3}
\end{figure}

\begin{figure}[h]
\captionsetup{font=footnotesize}
\centering
\includegraphics[width=\textwidth]{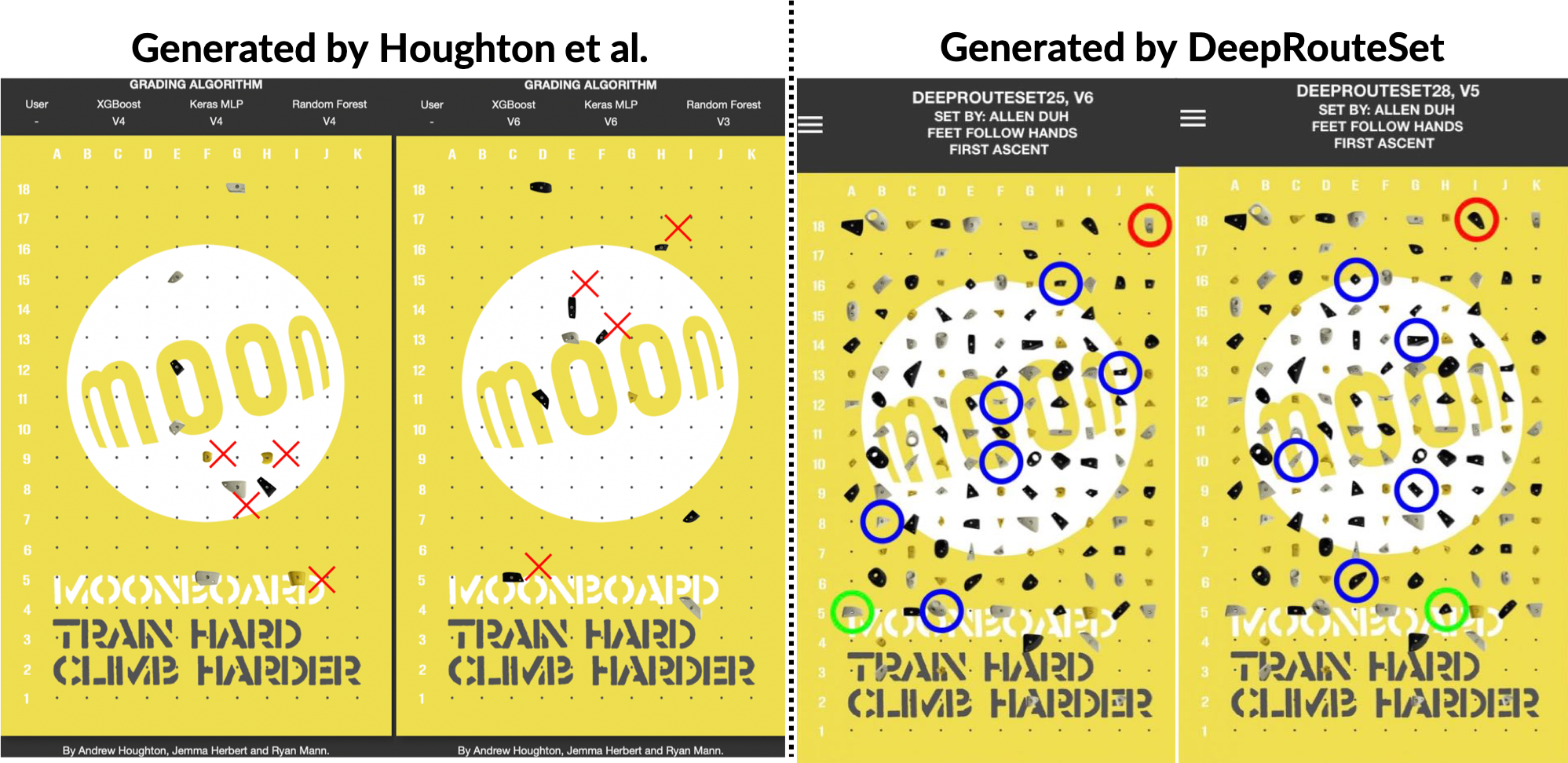}
\caption{Comparison between the generated problems by Houghton et al and DeepRouteSet. The holds that are labeled with “X” are the ones that are redundant.}
\label{fig:figA4}
\end{figure}

\end{document}